\documentclass[a4paper, 10 pt, conference]{hsmr} 
\usepackage[utf8]{inputenc}
\IEEEoverridecommandlockouts                       
\usepackage{mathtools}
\usepackage{geometry}
\usepackage{ifpdf}
\usepackage[labelsep=space, compatibility=false]{caption}
\usepackage{newtxmath,newtxtext}
\usepackage{authblk}
\usepackage{pgfplots}
\usepackage{graphicx}
\usepackage[colorlinks,urlcolor=blue]{hyperref} 
\usepackage{newtxtext,newtxmath}
\usepackage{makecell}
\usepackage{multirow}

\usepackage{graphicx}

\pgfplotsset{compat=newest} 

\title{\LARGE \bf
Gravity Compensation of the dVRK-Si Patient Side Manipulator
based on Dynamic Model Identification} 

\author{\LARGE Haoying Zhou$^{1,2}$, Hao Yang$^{3}$, Anton Deguet$^{2}$, Loris Fichera$^{1}$,\\ Jie Ying Wu$^{3}$ and Peter Kazanzides$^{2,4}$
\thanks{Corresponding author email: \textbf{\textit{hzhou6}@wpi.edu}. This work was supported by NSF AccelNet awards OISE-1927275 and OISE-1927354.}
}
\affil{\Large\textit{$^{1}$Department of Robotics Engineering, Worcester Polytechnic Institute, MA, USA}\\ 
\Large\textit{$^{2}$Laboratory for Computational Sensing and Robotics, JHU, MD, USA}\\ 
\Large\textit{$^{3}$Department of Computer Science, Vanderbilt University, TN, USA}\\ 
\Large\textit{$^{4}$Department of Computer Science, Johns Hopkins University, MD, USA}}
\begin{document}

\maketitle
\thispagestyle{empty}
\pagestyle{empty}

\section*{INTRODUCTION}

\begin{figure}[ht]
    \centering
    \includegraphics[width=0.4\linewidth]{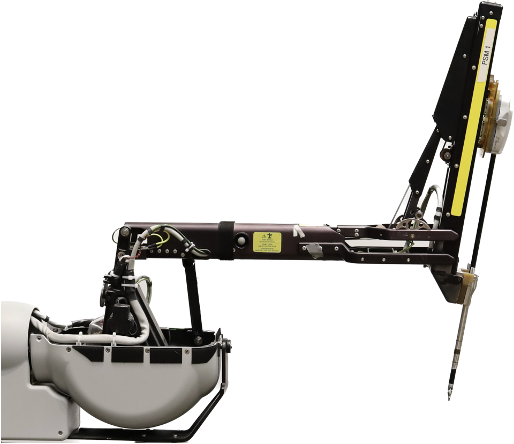}
    \includegraphics[width=0.4\linewidth]{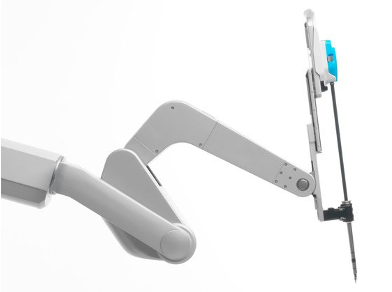}
    \caption{PSM arms. Left: dVRK Classic. Right: dVRK-Si.}
    \label{fig:psm_diff}
\end{figure}

The da Vinci Research Kit (dVRK, also known as dVRK Classic) is an open-source teleoperated surgical robotic system whose hardware is obtained from the first generation da Vinci Surgical System (Intuitive, Sunnyvale, CA, USA)~\cite{kazanzides2014open}. The dVRK has greatly facilitated research in robot-assisted surgery over the past decade and helped researchers address multiple major challenges in this domain~\cite{d2021accelerating}. Recently, the dVRK-Si system \cite{Xu2025}, a new version of the dVRK which uses mechanical components from the da Vinci Si Surgical System, became available to the community.
The major difference between the first generation da Vinci and the da Vinci Si is
in the structural upgrade of the Patient Side Manipulator (PSM), as shown in Fig~\ref{fig:psm_diff}. Because of this upgrade, the gravity of the dVRK-Si PSM can no longer be ignored as in the dVRK Classic. The high gravity offset may lead to relatively low control accuracy and longer response time~\cite{yang2024effectiveness}. In addition, although substantial progress has been made in addressing the dynamic model identification problem for the dVRK Classic~\cite{fontanelli2017modelling, wang2019convex, yang2024hybrid}, further research is required on model-based control for the dVRK-Si, due to differences in mechanical components and the demand for enhanced control performance. To address these problems, in this work, we present (1) a novel full kinematic model of the dVRK-Si PSM, and (2) a gravity compensation approach based on the dynamic model identification. 

\section*{MATERIALS AND METHODS}

The equations of motion for a robotic manipulator can be expressed using the well-known relation:
\begin{equation}
    \begin{split}
        M(q) \Ddot{q} + C(q, \Dot{q}) \Dot{q} + G(q) = \tau
    \end{split}
    \label{eq:robotdyn}
\end{equation}
where $q$ are the joint variables, 
$\tau$ are the joint torques.

To address the gravity compensation problem, we propose an approach where we identify the full dynamic model parameters and only take the $G(q)$ term as the force/torque required for gravity compensation.
Our proposed approach is as follows: (1) Build the kinematic model of the dVRK-Si PSM based on the parameters in Table~\ref{tab:psm_para}; (2) Construct the dynamic model of the PSM using the Euler-Lagrangian approach; (3) Calculate the optimal excitation trajectory based on the given joint constraints and the dynamic model~\cite{wang2019convex, yang2024hybrid}; (4) Run the excitation trajectory on the physical dVRK-Si PSM and collect kinematic \& dynamic data; (5) Solve for the model parameters using a convex optimization approach and enable the gravity compensation.
%
\subsection*{Kinematic Model}
\begin{figure}[ht]
    \centering
    \includegraphics[width=0.85\linewidth]{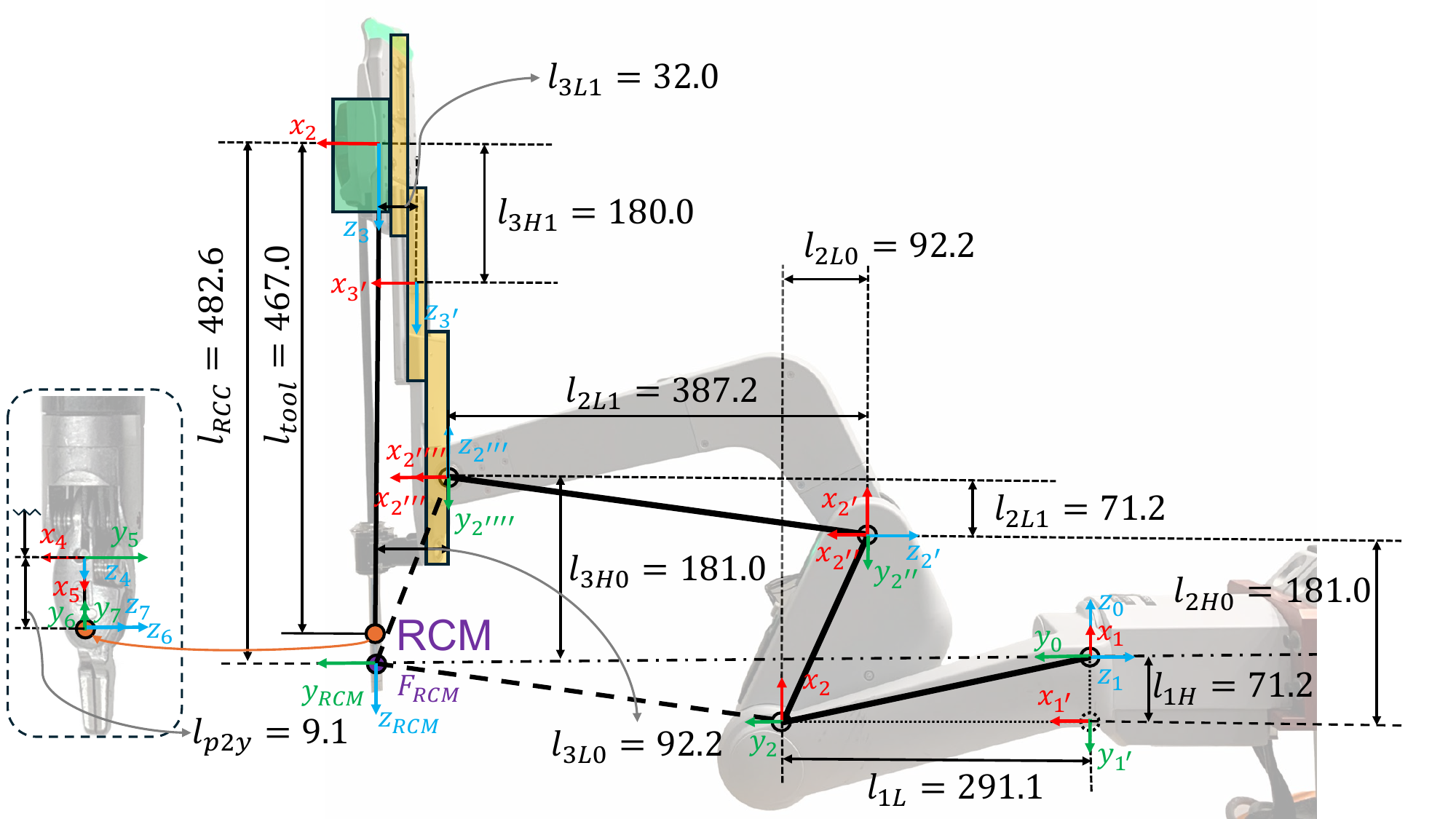}
    \caption{The planar view of the frame definition for the dVRK-Si PSM. The unit of the link lengths is mm. All frames are left-hand frames.}
    \label{fig:kinematics}
\end{figure}

Based on our own measurements and the specifications available from the manufacturer, we propose the full kinematic model of the dVRK-Si PSM as shown in Table~\ref{tab:psm_para} and Fig~\ref{fig:kinematics}.




\begin{table}[]
\caption{dVRK-Si PSM Kinematic Parameters}
\resizebox{\columnwidth}{!}{%
\begin{tabular}{llllllllll}
$i$      & Ref   & $a_{i-1}$ & $\alpha_{i-1}$   & $d_i$                    & $\theta_i$               & $\delta_{Li}$ & $I_{mi}$     & $F_i$        & $K_{si}$       \\
\hline
1        & 0     & 0         & $\frac{\pi}{2}$  & 0                        & $q_1+\frac{\pi}{2}$      & $\checkmark$  & $\times$     & $\checkmark$ & $\times$     \\
1'       & 1     & $-l_{1H}$ & $-\frac{\pi}{2}$ & 0                        & $\frac{\pi}{2}$          & $\times$      & $\times$     & $\times$     & $\times$     \\
2        & 1'    & $l_{1L}$  & 0                & 0                        & $q_2-\frac{\pi}{2}$      & $\checkmark$  & $\times$     & $\checkmark$ & $\times$     \\
2'       & 2     & $l_{2L0}$ & $\frac{\pi}{2}$  & $l_{2H0}$                & 0                        & $\times$      & $\times$     & $\times$     & $\times$     \\
2''      & 2'    & 0         & $-\frac{\pi}{2}$ & 0                        & $- q_2 + \frac{\pi}{2}$  & $\checkmark$  & $\times$     & $\checkmark$ & $\times$     \\
2'''     & 2''   & $l_{2L1}$ & $\frac{\pi}{2}$  & $l_{2H1}$                & 0                        & $\times$      & $\times$     & $\times$     & $\times$     \\
2''''    & 2'''  & 0         & $-\frac{\pi}{2}$ & 0                        & $q_2$                    & $\checkmark$  & $\times$     & $\checkmark$ & $\times$     \\
3        & 2'''' & $l_{2L2}$ & $-\frac{\pi}{2}$ & $q_3+l_{c2}$  & 0                        & $\checkmark$  & $\times$     & $\checkmark$ & $\times$     \\
3'       & 3     & $-l_{3L}$ & 0                & $l_{3H} - \frac{q_3}{2}$ & 0                        & $\times$      & $\times$     & $\checkmark$ & $\times$     \\
4        & 3     & 0         & 0                & $l_{tool}$               & $q_4$                    & $\times$      & $\checkmark$ & $\checkmark$ & $\checkmark$ \\
5        & 4     & 0         & $\frac{\pi}{2}$  & 0                        & $q_{5}+ \frac{\pi}{2}$ & $\times$      & $\checkmark$ & $\checkmark$ & $\times$     \\
6        & 5     & $l_{p2y}$ & $-\frac{\pi}{2}$ & 0                        & $q_{6}+ \frac{\pi}{2}$ & $\times$      & $\times$     & $\checkmark$ & $\times$     \\
7        & 5     & $l_{p2y}$ & $-\frac{\pi}{2}$ & 0                        & $q_{7}+ \frac{\pi}{2}$ & $\times$      & $\times$     & $\checkmark$ & $\times$     \\
$M_6$    & -     & 0         & 0                & 0                        & $q^m_6$                    & $\times$      & $\checkmark$ & $\checkmark$ & $\times$     \\
$M_7$    & -     & 0         & 0                & 0                        & $q^m_7$                    & $\times$      & $\checkmark$ & $\checkmark$ & $\times$     \\
$F_{67}$ & -     & 0         & 0                & 0                        & $q_{7}-q_{6}$        & $\times$      & $\times$     & $\checkmark$ & $\times$    
\end{tabular}%
}
\footnotesize{Note for the table: (1) Ref stands for the reference frame of link $i$. (2) $a_{i-1}$, $\alpha_{i-1}$, $d_i$ and $\theta_i$ are the modified DH parameters of link $i$. (3) $\delta_{Li}$, $I_{mi}$, $F_i$ and $K_{si}$ are the parameters of link inertia, motor inertia, joint friction, and stiffness of link $i$, respectively. (4) $M_6$ and $M_7$ represent the motor model of joint 6 and 7. $F_{67}$ represents the relative motion model between link 6 and 7. $q^m_i$ represents the equivalent motor movements at joint $i$.  (5) $l_{c2}=l_{2H1} - l_{RCC}$.}
\label{tab:psm_para}
\end{table}

\subsection*{Dynamic Parameter Identification}


The Euler-Lagrange method is used to calculate the dynamic parameters, 
$\delta$, as in~\cite{yang2024hybrid}. These are a combination of the first moment of inertia, friction, motor inertia and spring stiffness. To identify $\delta$, we apply linear parametrization to Eq.~\eqref{eq:robotdyn} to obtain $\tau = H(q, \Dot{q}, \Ddot{q})\delta$.

Therefore, the identification problem can be converted into an optimization problem which minimizes a cost function $J_c$, where $n$ is the number of data points:
\begin{equation}
    \begin{split}
        J_c &= ||\mathbf{W}\delta - \mathbf{T}||^2 \\ 
        \mathbf{W} &= \begin{bmatrix} H(q_1, \Dot{q}_1, \Ddot{q}_1), \hdots, H(q_n, \Dot{q}_n, \Ddot{q}_n) \end{bmatrix}^T \\ \mathbf{T} &= \begin{bmatrix} \tau_1, \tau_2, \hdots, \tau_n \end{bmatrix}^T
    \end{split}
\end{equation}

To reduce overfitting, we utilize physical consistency constraints when solving for the dynamic parameters: (1) The mass of each link $k$ is positive. (2) The center of mass for the link stays in its convex hull. (3) The friction coefficients, motor inertia and spring stiffness are positive. 

\section*{RESULTS}

To validate our gravity compensation results, we perform a drift test by first moving the PSM to a desired pose using its position controller (a PD controller) and then switching to the (open loop) gravity compensation (GC) torque-based control model for 5 seconds. The robot is considered to be drifting when moving more than 1 degree or 1 mm within the 5 seconds. We collect the kinematic data of the PSM and report the errors between the desired joint values and the measured joint values. We also determine the lower/upper bound torques (LB/UB $\tau$) which, due to static friction, are sufficient to hold the joint in position.  From Table~\ref{tab:psm_para}, we can see that gravity mainly influences the first three joints, and thus we only focus on those measurements and evaluations. We select five random poses for testing and the results are shown in Table \ref{tab:drift_test}.

\begin{table}[htbp]
    \scriptsize
    \begin{center}
    \caption{Drift Test Results}
    \begin{tabular}{|c|c|c|c|c|c|c|}
        \hline
        \makecell{\textbf{Item}} & \makecell{\textbf{PD} \\ \textbf{Pos} \\ \textbf{Err}} & \makecell{\textbf{GC} \\ \textbf{Pos} \\ \textbf{Err}} & \makecell{\textbf{PD} \\ $\mathbf{\tau}$} & \makecell{\textbf{GC} \\ $\mathbf{\tau}$} & \makecell{\textbf{Non-} \\ \textbf{Drift} \\ \textbf{LB} \\ $\mathbf{\tau}$} & \makecell{\textbf{Non-} \\ \textbf{Drift} \\ \textbf{UB} \\ $\mathbf{\tau}$} \\
        \hline
        \multirow{5}{*}{\makecell{Joint1 \\ (deg \\or\\N$\cdot$m)}} & 0.14 & 0.04 & -1.42 & -1.07 & -0.11 & -1.76\\
        \cline{2-7}
        & 0.06 & 0.04 & 0.60 & 1.25 & 0.43 & 2.17\\
        \cline{2-7}
        & 0.85 & 0.16 & 8.87 & 7.67 & 6.06 & 9.75\\
        \cline{2-7}
        & 0.50 & 0.11 & -5.25 & -6.39 & -4.66 & -7.99\\
        \cline{2-7}
        & 0.27 & 0.05 & -2.81 & -3.61 & -2.48 & -4.51\\
        \hline
        \multirow{5}{*}{\makecell{Joint2 \\ (deg \\or\\N$\cdot$m)}} & 0.50 & 0.13 & 5.29 & 4.32 & 2.98 & 5.75 \\
        \cline{2-7}
        & 0.45 & 0.12 & 4.67 & 5.73 & 3.84 & 7.34\\
        \cline{2-7}
        & 0.03 & 0.03 & 0.31 & 0.08 & -0.87 & 0.85\\
        \cline{2-7}
        & 0.03 & 0.02 & -0.32 & -1.44 & -0.18 & -2.64\\
        \cline{2-7}
        & 0.37 & 0.14 & -3.83 & -2.97 & -1.96 & -4.61\\
        \hline
        \multirow{5}{*}{\makecell{Joint3 \\ (mm \\or\\N)}} & 1.49 & 0.25 & -8.95 & -7.79 & -4.75 & -9.35\\
        \cline{2-7}
        & 0.36 & 0.13 & -2.19 & -4.41 & -2.07 & -5.96\\
        \cline{2-7}
        & 0.46 & 0.63 & -2.77 & 0.14 & -2.79 & 2.03\\
        \cline{2-7}
        & 1.05 & 0.35 & -6.31 & -4.56 & -1.83 & -6.67\\
        \cline{2-7}
        & 0.78 & 0.39 & 4.68 & 2.97 & 0.92 & 4.78\\
        \hline
        \hline
        \textbf{Pose} & \textbf{x} & \textbf{y} & \textbf{z} & $\mathbf{r_x}$ & $\mathbf{r_{y}}$ & $\mathbf{r_{z}}$\\
        \hline
        1 & 0.00 & 0 .00 & 113.50 & 180.00 & 0.00 & -90.00 \\
        \hline
        2 & -123.93 & 101.14 & 108.87 & -126.83 & -20.18 & -65.27\\
        \hline
        3 & -78.68 & -50.46 & -2.30 & -88.59 & -0.90 & -122.65\\
        \hline
        4 & 50.10 & -45.76 & 48.66 & 129.09 & 22.45 & -64.83\\
        \hline
        5 & 174.45 & 67.87 & -79.84 & 66.66 & 7.98 & -79.84\\
        \hline
    \end{tabular}
    \label{tab:drift_test}
    \end{center}
    \footnotesize{Note for the table: The units of the poses are mm and deg. The 6D pose coordinates are based on the frame $F_{RCM}$ shown in Fig.~\ref{fig:kinematics}.}
\end{table}

A video of the free space movements with gravity compensation enabled is shown in \url{https://github.com/jackzhy96/dVRK_Si_PSM_gravity_compensation}.

\section*{DISCUSSION}

Results indicate that the GC torque-based control effectively maintains the robot's pose without noticeable drift. Furthermore, the GC approach usually achieves a smaller steady-state error compared to the PD control. The joint torques generated by the GC method for all three measured joints fall within their respective lower and upper bounds and, in most cases, are close to the midpoint of that range. For the third joint, some offset from the midpoint can be attributed to the significant friction in the prismatic joint and the cannula. Preliminary investigations also show that the gravity compensation can be generalized on different dVRK-Si PSMs with the same mounting-angle configuration. 


\nocite{*}
\bibliographystyle{IEEEtran}
\bibliography{references}

\begin{thebibliography}{1}
\providecommand{\url}[1]{#1}
\csname url@samestyle\endcsname
\providecommand{\newblock}{\relax}
\providecommand{\bibinfo}[2]{#2}
\providecommand{\BIBentrySTDinterwordspacing}{\spaceskip=0pt\relax}
\providecommand{\BIBentryALTinterwordstretchfactor}{4}
\providecommand{\BIBentryALTinterwordspacing}{\spaceskip=\fontdimen2\font plus
\BIBentryALTinterwordstretchfactor\fontdimen3\font minus \fontdimen4\font\relax}
\providecommand{\BIBforeignlanguage}[2]{{%
\expandafter\ifx\csname l@#1\endcsname\relax
\typeout{** WARNING: IEEEtran.bst: No hyphenation pattern has been}%
\typeout{** loaded for the language `#1'. Using the pattern for}%
\typeout{** the default language instead.}%
\else
\language=\csname l@#1\endcsname
\fi
#2}}
\providecommand{\BIBdecl}{\relax}
\BIBdecl

\bibitem{kazanzides2014open}
P.~Kazanzides, Z.~Chen, A.~Deguet, G.~S. Fischer, R.~H. Taylor, and S.~P. DiMaio, ``An open-source research kit for the {da Vinci{\textregistered} Surgical System},'' in \emph{IEEE Intl. Conf. on Robotics and Automation (ICRA)}, 2014, pp. 6434--6439.

\bibitem{d2021accelerating}
C.~D’Ettorre, A.~Mariani, A.~Stilli, F.~R. y~Baena, P.~Valdastri, A.~Deguet, P.~Kazanzides, R.~H. Taylor, G.~S. Fischer, S.~P. DiMaio \emph{et~al.}, ``Accelerating surgical robotics research: A review of 10 years with the {da Vinci Research Kit},'' \emph{IEEE Robotics \& Automation Magazine}, vol.~28, no.~4, pp. 56--78, 2021.

\bibitem{Xu2025}
K.~Xu, J.~Y. Wu, A.~Deguet, and P.~Kazanzides, ``{dVRK-Si}: The next generation {da Vinci Research Kit},'' in \emph{IEEE Intl. Symp. of Medical Robotics (ISMR)}, 2025, (in review).

\bibitem{yang2024effectiveness}
H.~Yang, A.~Acar, K.~Xu, A.~Deguet, P.~Kazanzides, and J.~Y. Wu, ``An effectiveness study across baseline and neural network-based force estimation methods on the {da Vinci Research Kit Si System},'' in \emph{arXiv:2405.07453 [cs.RO]}, 2024.

\bibitem{fontanelli2017modelling}
G.~A. Fontanelli, F.~Ficuciello, L.~Villani, and B.~Siciliano, ``Modelling and identification of the {da Vinci Research Kit} robotic arms,'' in \emph{IEEE/RSJ Intl. Conf. on Intelligent Robots and Systems (IROS)}, 2017, pp. 1464--1469.

\bibitem{wang2019convex}
Y.~Wang, R.~Gondokaryono, A.~Munawar, and G.~S. Fischer, ``A convex optimization-based dynamic model identification package for the {da Vinci Research Kit},'' \emph{IEEE Robotics and Automation Letters}, vol.~4, no.~4, pp. 3657--3664, 2019.

\bibitem{yang2024hybrid}
H.~Yang, H.~Zhou, G.~S. Fischer, and J.~Y. Wu, ``A hybrid model and learning-based force estimation framework for surgical robots,'' in \emph{IEEE/RSJ Intl. Conf. on Intelligent Robots and Systems (IROS)}, 2024, pp. 906--912.

\end{thebibliography}

\end{document}